\documentclass{article}

\usepackage{paralist}
\usepackage{amsmath}
\usepackage{amssymb}
\usepackage{amsthm}
\usepackage{algorithmic}

\usepackage{microtype}
\usepackage{graphicx}
\usepackage{subfigure}
\usepackage{booktabs} 

\usepackage{hyperref}

\usepackage{enumitem}

\newcommand{\bx}{{\bf x}}
\newcommand{\by}{{\bf y}}

\newcommand{\bw}{{\bf w}}
\newcommand{\balpha}{{\bf \alpha}}
\newcommand{\sY}{\mathcal{Y}}

\newcommand{\lbl}[1]{{\bf \tt #1}}

\newcommand{\p}[1]{\left(#1\right)}
\newcommand{\f}[2]{\Phi\p{#1,#2}}
\newcommand{\fk}[3]{\Phi_{#1}\p{#2,#3}}

\newtheorem{theorem}{Theorem}

\newtheorem{lemma}{Lemma}

\usepackage[accepted]{icml2018}

\newcommand{\theTitle}{Learning to Speed Up Structured Output Prediction}

\icmltitlerunning{\theTitle}

\begin{document}

\twocolumn[
\icmltitle{\theTitle}




\begin{icmlauthorlist}
\icmlauthor{Xingyuan Pan}{ut}
\icmlauthor{Vivek Srikumar}{ut}
\end{icmlauthorlist}

\icmlaffiliation{ut}{School of Computing, University of Utah, Salt Lake City, Utah, USA}

\icmlcorrespondingauthor{Xingyuan Pan}{xpan@cs.utah.edu}
\icmlcorrespondingauthor{Vivek Srikumar}{svivek@cs.utah.edu}

\icmlkeywords{Structured Prediction, Learning to Search}

\vskip 0.3in
]



\printAffiliationsAndNotice{}  

\begin{abstract}
  Predicting structured outputs can be computationally onerous due to
  the combinatorially large output spaces.  In this paper,
  we focus on reducing the prediction time of a trained black-box
  structured classifier without losing accuracy.
  To do so, we train a {\em speedup} classifier that learns to mimic a
  black-box classifier under the learning-to-search approach.
  As the structured classifier predicts more examples, the
  speedup classifier will operate as a learned heuristic to guide
  search to favorable regions of the output space.
  We present a mistake bound for the speedup classifier and identify
  inference situations where it can independently make correct
  judgments without input features.
  We evaluate our method on the task of entity and relation extraction
  and show that the speedup classifier outperforms even greedy search in terms
  of speed without loss of accuracy.
\end{abstract}



\section{Introduction}
\label{sec:intro}

Many natural language processing (NLP) and computer vision problems
necessitate predicting
structured outputs such as labeled sequences, trees or general
graphs~\cite{smith2010linguistic,nowozin2011structured}.
Such tasks require modeling both  input-output relationships and 
the interactions between predicted outputs to capture correlations.
Across the various structured prediction
formulations~\cite{Lafferty2001,Taskar2003a,chang2012structured},
prediction requires solving inference problems by searching for
score-maximizing output structures. The search space for inference
is typically large (e.g., all parse trees), and grows with input
size.  Exhaustive search can be prohibitive and standard
alternatives are either:
\begin{inparaenum}[(a)]
\item perform exact inference with a large computational cost
  or,
\item approximate inference to sacrifice accuracy in favor of time.
\end{inparaenum}

In this paper, we focus on the computational cost of inference.  We
argue that naturally occurring problems have remarkable
regularities across both inputs and outputs, and traditional
formulations of inference ignore them. For example, parsing an
$n$-word sentence will cost a standard head-driven lexical parser
$O(n^5)$ time. Current practice in NLP is to treat each new sentence
as a fresh discrete optimization problem and pay the computational
price each time.  However, this practice is not only expensive, but
also wasteful!  We ignore the fact that slight changes to
inputs often do not change the output, or even the sequence of steps
taken to produce it. Moreover, not all outputs are linguistically
meaningful structures; as we make more predictions, we should be
able to {\em learn} to prune the output space.

The motivating question that drives our work is: {\em Can we design
  inference schemes that learn to make a trained structured
  predictor faster without sacrificing output quality?}  After
training, the structured classifier can be thought as a
black-box. Typically, once deployed, it is never modified over its
lifetime of classifying new examples. Subsequently, we can view each
prediction of the black-box classifier as an opportunity to learn
how to navigate the output space more efficiently. Thus, if the
classifier sees a previously encountered situation, it could make
some decisions without needless computations.


We formalize this intuition by considering the trained models as
solving arbitrary integer linear programs (ILPs) for combinatorial
inference. We train a second, inexpensive {\em speedup} classifier
which acts as a heuristic for a search-based inference algorithm
that mimics the more expensive black-box classifier. The speedup
heuristic is a function that learns regularities among {\em
  predicted} structures.  We present a mistake bound algorithm that,
over the classifier's lifetime, learns to navigate the feasible
regions of the ILPs. By doing so, we can achieve a reduction in
inference time.

We further identify inference situations where the learned speedup
heuristic alone can correctly label parts of the outputs without
computing the corresponding input features. In such situations, the
search algorithm can safely {\em ignore} parts of inputs if the
corresponding outputs can be decided based on the sub-structures
constructed so far. Seen this way, the speedup classifier can be
seen as a statistical cache of past decisions made by the black-box
classifier.

We instantiate our strategy to the task of predicting entities and
relations from sentences. Using an ILP based black-box classifier,
we show that the trained speedup classifier mimics the reference
inference algorithm to obtain improvements in running time, and also
recovers its accuracy. Indeed, by learning to ignore input
components when they will not change the prediction, we show that
learned search strategy outperforms even greedy search in terms of
speed.

To summarize, the main contribution of this paper is the formalization
of the problem of learning to make structured output classifiers
faster without sacrificing accuracy. We develop a learning-to-search
framework to train a speedup classifier with a mistake-bound guarantee
and a sufficient condition to safely avoid computing input-based
features. We show empirically on an entity-relation extraction task
that we can learn a speedup classifier that is (a) faster than both
the state-of-the-art Gurobi optimizer and greedy search, and (b) does
not incur a loss in output quality.


\section{Notation and Preliminaries}
\label{sec:notation}

First, we will define the notation used in this paper with a
running example that requires of identifying entity types
and their relationships in text. The input to the problem consists of
sentences such as:
\vspace{-0.1in}
\begin{enumerate}
  \item[] \underline{Colin} went back home in \underline{Ordon Village}.
\end{enumerate}
\vspace{-0.1in}
These inputs are typically preprocessed --- here, we are given spans of
text (underlined) corresponding to entities. We will denote such
preprocessed inputs to the structured prediction problem as $\bx$.  

\begin{figure}[tbp] 
   \includegraphics[width=3in]{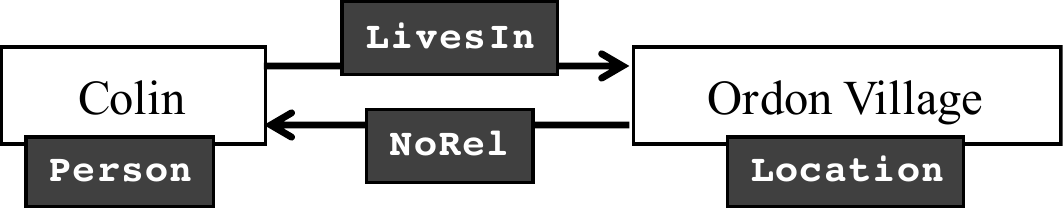} 
   \caption{An example of the entities and relations task. The nodes
     are entity candidates and directed edges indicate relations. The
     labels in typewriter font are the decisions that we need to
     make.}
   \label{fig:er-example}
\end{figure}

We seek to produce a structure $\by \in \sY_\bx$ (e.g., labeled trees,
graphs) associated with these inputs. Here, $\sY_\bx$ is the set of
all possible output structures for the input $\bx$. In the example
problem, our goal is to assign types to the entities and also label
the relationships between them. Suppose our task has three types of
entities: \lbl{person}, \lbl{location} and
\lbl{organization}.  A pair of entities can participate in one of five
possible directed relations: \lbl{Kill}, \lbl{LiveIn}, \lbl{WorkFor},
\lbl{LocatedAt} and \lbl{OrgBasedIn}. Additionally, there is a special
entity label \lbl{NoEnt} meaning a text span is not an entity, and a
special relation label \lbl{NoRel} indicating that two spans are unrelated.
Figure~\ref{fig:er-example} shows a plausible
structure for the example sentence as per this scheme.

A standard way to model the prediction problem requires learning a
model that scores all structures in $\sY_\bx$ and searching for the
score-maximizing structure. Linear models are commonly used as
scoring functions, and require a feature vector characterizing
input-output relationships $\f{\bx}{\by}$. We will represent the
model by a weight vector $\balpha$. Every structure $\by$ associated
with an input $\bx$ is scored as the dot product $\balpha \cdot \f{\bx}{\by}$.
The goal of
prediction is to find the structure $\by^*$ that maximizes this
score. That is,
\begin{equation}\label{general-argmax}
\by^* = \text{arg}\max_{\by\in\sY_\bx} \alpha \cdot \f{\bx}{\by}.
\end{equation} 
Learning involves using training data to find the best weight vector $\balpha$.

In general, the output structure $\by$ is a set of
$K$ categorical inference variables $\{y^1, y^2, \cdots, y^K\}$ , each
of which can take a value from a predefined set of $n$ labels. That
is, each $y^k \in \by$ takes a value from $\{l_1, l_2, \cdots, l_n\}$.\footnote{We make this choice for simplicity of notation. In general, $K$ depends on the size of the input $\bx$, and categorical variables may take values from different label sets.}  In our running example,
the inference variables correspond to the four decisions
that define the structure: the labels for the two entities, and the
relations in each direction.
The feature function $\Phi$ decomposes into a sum of features over
each $y^k$, each denoted by $\Phi_k$, giving us the inference problem:
\begin{equation}
  \label{argmax}
  \by^* = \text{arg}\max_{\by\in\sY_\bx} \sum_{k=1}^K\alpha \cdot \fk{k}{\bx}{y^k}.  
\end{equation}
The dependencies between the $y^k$'s specify the nature of the output
space.
Determining each $y^k$ in isolation greedily does not typically
represent a viable inference strategy because constraints connecting
the variables are ignored. In this spirit, the problem of finding the
best structure can be viewed as a combinatorial optimization
problem.

{\em In this paper, we consider the scenario in which we have already
  trained a model $\alpha$.  We focus on solving the inference problem
  (i.e.,Eq.~\eqref{argmax}) efficiently.}  We conjecture
that it should be possible to observe a black-box inference algorithm
over its lifetime to learn to predict faster without losing accuracy.


\subsection{Black-box Inference Mechanisms}
\label{sec:ilp}

One common way to solve inference is by designing efficient dynamic
programming algorithms that exploit problem structure.
While effective, this approach is limited to special cases where the
problem admits efficient decoding, thus placing restrictions
on factorization and feature design.

In this paper, we seek to reason about the problem of predicting structures in the
general case. Since inference is essentially a
combinatorial optimization problem, without loss of generality, we can
represent any inference problem as an integer linear programming (ILP)
instance~\cite{Schrijver1998}. To represent the inference task in
Eq.~\eqref{argmax} as an ILP instance, we will define indicator
variables of the form $z_i^k \in\{0,1\}$, which stands for the
decision that the categorical variable $y^k$ is assigned the $i^{th}$
label among the $n$ labels.That is, $z^k_i=1$ if $y^k=l_i$, and $0$ otherwise. Using this notation, we can write the
cost of any structure $\by$ in terms of the indicators as
\begin{equation}
\sum_{k=1}^K \sum_{i=1}^n c^k_i z^k_i. \label{primal}
\end{equation}
Here, $c^k_i$ is a stand in for
$-\alpha \cdot \fk{k}{\bx}{l_i}$, namely the cost (negative score) associated
with this decision.\footnote{The negation defines an equivalent
minimization problem and makes subsequent description of the search
framework easier.}
In our example, suppose the first categorical variable $y^1$
corresponds to the entity {\it Colin}, and it has possible labels
$\{\text{\lbl{person}}, \text{\lbl{location}}, \dots\}$. Then,
assigning \lbl{person} to {\it Colin} would correspond to
setting $z^1_1=1$, and $z^1_i=0$ for all $i\ne 1$. Using the labels
enumerated in \S\ref{sec:notation}, there will be 20 indicators for
the four categorical decisions.

Of course, arbitrary assignments to the indicators is not allowed. We
can define the set of feasible structures using linear
constraints. Clearly, each categorical variable can take 
exactly one label, which can be expressed via:
\begin{equation}
  \sum_{i=1}^n z^k_i = 1, \quad \text{for all } k. \label{unique}
\end{equation}
In addition, we can define the set of valid structures
$\sY_\bx$ using a collection of $m$ linear constraints, the $j^{th}$ one
of which can be written as
\begin{equation}
\sum_{k=1}^K \sum_{i=1}^n A^k_{ji} z^k_i = b_j, \quad \text{for all } j. \label{structure}
\end{equation}
These {\em structural} constraints characterize the interactions between the
categorical variables. For example, if a directed edge in our running
example is labeled as \lbl{LiveIn}, then, its source  and target must be a
\lbl{person} and a \lbl{location} respectively. While
Eq.\eqref{structure} only shows equality constraints, in practice,
inequality constraints can also be included.

The inference problem in Eq.~\eqref{argmax} is equivalent to the
problem of minimizing the objective in Eq.~\eqref{primal} over the
$0$-$1$ indicator variables subject to the constraints in
Eqs.~\eqref{unique} and \eqref{structure}. 

We should note the difference between the ability to write an
inference problem as an ILP instance and actually {\em solving} it
as one. The former gives us the ability to reason about inference in
general, and perhaps using other methods (such as Lagrangian
relaxation~\cite{Lemarechal2001}) for inference.  However, solving
problems with industrial strength ILP solvers such as the Gurobi
solver\footnote{\url{http://www.gurobi.com}} is competitive with
other approaches in terms of inference time, even though they may not
directly exploit problem structure.

In this work, we use the general structure of the ILP inference
formulation to develop the theory for speeding up inference. In
addition, because of its general applicability and fast inference
speed, we use the Gurobi ILP solver as our black-box classifier, and
learn a speedup heuristic to make even faster inference.



\subsection{Inference as Search}
\label{sec:search}

Directly applying the black-box solver for the large output spaces may
be impractical. An alternative general purpose strategy for inference
involves framing the maximization in Eq.~\eqref{argmax} as a graph
search problem. 

Following \citet{russell2003artificial,Xu2009a}, a general graph
search problem requires defining an initial search node $I$, a
successor function $s(\cdot)$, and a goal test. The successor
function $s(\cdot)$ maps a search node to its successors. The goal
test determines whether a node is a goal node. Usually, each search
step is associated with a cost function, and we seek to find a goal
node with the least total cost.

We can define the search problem corresponding to inference as
follows. We will denote a generic search node in the graph as $v$, which
corresponds to a set of partially assigned categorical
variables. Specifically, we will define the search node $v$ as a set
of pairs $\{(k, i)\}$, each element of which specifies that the
variable $y^k$ is assigned the $i^{th}$ label.  The initial search
node $I$ is the empty set since none of the variables has been
assigned when the search begins. For a node $v$, its successors
$s(v)$ is a set of nodes, each containing one more assigned variable
than $v$. A node is a goal node if all variables $y^k$'s have been
assigned. The size of any goal node is $K$, the number of categorical
variables.

In our running example, at the start of search, we may choose to
assign the first label $l_1$ (\lbl{person}) to the variable $y^1$ -- the
entity {\em Colin} -- leading us to the successor
$\{(1,1)\}$. Every search node specifies a partial or a full
assignment to all the entities and relations. The goal test simply
checks if we arrive at a full assignment, i.e., all the entity and
relation candidates have been assigned a label. 

Note that goal test does not test the quality of the node, it simply
tests whether the search process is finished. The quality of the
goal node is determined by the path cost from the initial node to the
goal node, which is the accumulated cost of each step along the
way. The step cost for assigning label $l_i$ to a variable
$y^k$ is  the same $c^k_i$ we defined for the ILP objective
in Eq.~\eqref{primal}. Finding a shortest path in such
a search space is equivalent to the original ILP problem without the structural
constraints in Eq.~\eqref{structure}. The unique-label constraints in Eq.~\eqref{unique} are automatically satisfied by our formulation of the search process.

Indeed, solving inference without the constraints in
Eq.\eqref{structure} is trivial. For each categorical variable $y^k$,
we can pick the label $l_i$ that has the lowest value of $c^k_i$. This
gives us two possible options for solving inference as search: We can
\begin{inparaenum}[(a)]
\item ignore the constraints that make inference slow to greedily
  predict all the labels, or,
\item enforce constraints at each step of the search, and only
  consider search nodes that satisfy all constraints.
\end{inparaenum}
The first option is fast, but can give us outputs that are
invalid. For example, we might get a structure that mandates that the
\lbl{person} {\em Colin} lives in a \lbl{person} called {\em Ordon
  Village}. The second option will give us structurally valid outputs,
but can be prohibitively slow.

Various graph search algorithms can be used for performing
inference. For efficiency, we can use beam search with a fixed beam
width $b$. When search begins the beam $B_0$ contains only the initial
node $B_0 = [I]$. Following \citet{Collins2004,Xu2009a}, we define the
function \textbf{BreadthExpand} which takes the beam $B_t$ at step $t$
and generates the candidates $C_{t+1}$ for the next beam:
\begin{align*}
C_{t+1} &= \text{\textbf{BreadthExpand}}(B_t)\\& = \cup_{v\in B_t}s(v)
\end{align*}
The next beam is given by
$B_{t+1}=\text{\textbf{Filter}}(C_{t+1})$, where \textbf{Filter} takes
top $b$ nodes according to some priority function $p(v)$. In the
simplest case, the priority of a node $v$ is the total path cost of
reaching that node. More generally, the priority function can be
informed not only by the path cost, but also by a heuristic
function as in the popular $A^*$ algorithm.


\section{Speeding up Structured Prediction}
\label{sec:speedup}

In the previous section, we saw that using a black-box ILP solver may
be slower than greedy search which ignores constraints, but produces valid outputs. 
However, over its lifetime, a trained classifier predicts structures
for a large number of inputs. While the number of unique inputs
(e.g. sentences) may be large, the number of unique structures that
actually occur among the predictions is not only finite, but also
small. This observation was exploited by \citet{Srikumar2012a,Kundu2013} for
amortizing inference costs. 

In this paper, we are driven by the need for an inference algorithm
that learns regularities across outputs to become faster at producing
structurally valid outputs. In order to do so, we will develop an
inference-as-search scheme that inherits the speed of greedy search,
but learns to produce structurally valid outputs.  Before developing
the algorithmic aspects of such an inference scheme, let us first see
a proof-of-concept for such a scheme.

\subsection{Heuristics for Structural Validity}
\label{sec:heuristics-validity}

Our goal is to incorporate the structural constraints from
Eq.~\eqref{structure} as a heuristic for greedy or beam search. To do
so, at each step during search, we need to estimate how likely an
assignment can lead to a constraint violation.  This information can
be characterized by using a heuristic function $h(v)$, which will be
used to evaluated a node $v$ during search.

The dual form the ILP in Eqs.~\eqref{primal} to \eqref{structure}
help justify the idea of capturing constraint information using a heuristic
function. We treat the unique label
constraints in Eq.~\eqref{unique} as defining the
domain in which each $0$-$1$ variable $z^k_i$ lives, and the only
real constraints are given by Eq.~\eqref{structure}.

Let $u_j$ represent the dual variable for the $j^{\text{th}}$ constraint. Thus, we
obtain the Lagrangian\footnote{We omit the ranges of the summation indices $i, j, k$ hereafter.}
\begin{align}
L(z,u) & =\sum_{k=1}^K \sum_{i=1}^n c^k_i z^k_i - \sum_{j=1}^m u_j\left(\sum_{k=1}^K\sum_{i=1}^n A^k_{ji} z^k_i-b_j\right) \nonumber \\
       & =\sum_{k,i}\left(c^k_i-\sum_j u_jA^k_{ji}\right)z^k_i + \sum_j b_j u_j \nonumber
\end{align}
The dual function $\theta(u) = \min_z L(z, u)$, where the minimization
is over the domain of the $z$ variables.

Denote $u^*=\text{arg}\max\theta(u)$ as the solution to the dual
problem. In the case of zero duality gap, the theory of Lagrangian
relaxation \cite{Lemarechal2001} tells us that solving the following relaxed
minimization problem will solve the original ILP:
\begin{align}
 & \min \sum_{k,i}\left(c^k_i-\sum_j u^*_jA^k_{ji}\right)z^k_i \label{dual-obj} \\
 & \sum_{i} z^k_i = 1, \quad \text{for all $k$}                                 \\
 & z^k_i \in \{0, 1\} , \quad \text{for all $k, i$} \label{01}
\end{align}
This new optimization problem does not have any structural constraints
and can be solved greedily for each $k$ if we know the optimal dual variables $u^*$.

To formulate the minimization in Eqs~\eqref{dual-obj} to \eqref{01} as a search problem, we define the priority function $p(v)$ for ranking the nodes as $p(v) = g(v) + h^*(v)$, where the path cost $g(v)$ and heuristic function $h^*(v)$ are given by
\begin{align}
g(v)&=\sum_{(k,i)\in v} c^k_i, \\
h^*(v)&= -\sum_{(k,i)\in v}\sum_j A^k_{ji} u^*_j(\bx). \label{h}
\end{align}
Since Eq.~\eqref{dual-obj} is a minimization problem, smaller priority value $p(v)$ means higher ranking during search.
Note that even though heuristic function defined in
this way is not always admissible, greedy search with ranking function $p(v)$ will lead to the exact solution of Eqs.~\eqref{dual-obj} to \eqref{01}. In practice, however, we do not
have the optimal values for the dual variables $u^*$. Indeed, when
Lagrangian relaxation is used for inference, the optmial dual
variables are computed using subgradient optimization for
each example because their value depends on the original input via
the $c$'s.

Instead of performing expensive gradient based optimization for {\em
  every} input instance, we will approximate the heuristic function as
a classifier that learns to prioritize structurally valid outputs. In
this paper, we use a linear model based on a weight vector $\bw$ to
approximate the heuristic as
\begin{equation}
h(v) = - \bw \cdot \phi(v)
\end{equation}
For an appropriate choice of node features
$\phi(v)$, the heuristic $h(v)$ in Eq.\eqref{h} is indeed a linear
function.\footnote{See supplementary material for an elaboration.} In other words, there exists a linear heuristic function
that can guide graph search towards creating structurally valid outputs.

In this setting, the priority function $p(v)$ for each node is
determined by two components: the path cost $g(v)$ from the initial
node to the current node, and the learned heuristic cost $h(v)$, which
is an estimate of how good the current node is.  Because the purpose
of the heuristic is to help improve inference speed, we call $\phi(v)$
speedup features. The speedup features can be different from the
original model features in Eq.~\eqref{argmax}. In particular it can
includes features for partial assignments made so far which were not
available in the original model features. In this setting, the goal of
speedup learning is to find suitable weight vector $\bf w$ over the
black-box classifier's lifetime.


\section{Learning the Speedup Classifier}
\label{sec:learning}
In this section, we will describe a mistake-bound algorithm to learn
the weight vector $\bw$ of the speedup classifier. The design of this
algorithm is influenced by learning to search algorithms such as
LaSO~\cite{DaumeIII2005,Xu2009a}.  We assume that we have access to a
trained black-box ILP solver called \textbf{Solve}, which can solve the
structured prediction problems, and we have a large set of examples
$\{\bx_i\}_{i=1}^N$. Our goal is to use this set to train a speedup
classifier to mimic the ILP solver while predicting structures for this set of
examples. Subsequently, we can use the less expensive speedup
influenced search procedure to replace the ILP solver.

To define the algorithm, we will need additional terminology.  Given a
reference solution $\by$, we define a node $v$ to be {\em $y$-good},
if it can possibly lead to the reference solution.
If a node $v$ is $y$-good, then the already assigned variables have
the same labels as in the reference solution. We define a beam $B$ is
$y$-good if it contains at least one $y$-good node to represent
the notion that search is still viable.
We denote the first element (the highest ranked) in a beam by
$\hat{v}$. 
Finally, we define an operator \textbf{SetGood}, which takes a node
that is not $y$-good, and return its corresponding $y$-good
node by fixing the incorrect assignments according to the reference solution. The unassigned
variables are still left unassigned by the \textbf{SetGood} operator.
\begin{algorithm}[t]
\caption{Learning a speedup classifier using examples $\{\bx_i\}_{i=1}^N$, and a black-box Solver \textbf{Solve}.}
\label{speedup}
\begin{algorithmic}[1]
\STATE Initialize the speedup weight vector ${\bf w} \leftarrow \bf 0$
\FOR{$epoch= 1 \dots M$}
	\FOR{$i= 1 \dots N$}
	\STATE $\by \leftarrow \text{\textbf{Solve}}(\bx_i)$
	\STATE Initialize the beam $B\leftarrow[I]$
	\WHILE{$B$ is $y$-good \AND $\hat{v}$ is not goal}
		\STATE $B \leftarrow \text{\textbf{Filter}}(\text{\textbf{BreadthExpand}}(B))$
	\ENDWHILE
	\IF{$B$ is not $\by$-good}
		\STATE $v^* \leftarrow \text{\textbf{SetGood}}(\hat{v})$
		\STATE $ \bw \leftarrow \bw + \phi(v^*) - \frac{1}{|B|}\sum_{v \in B} \phi(v)$
	\ELSIF{$\hat{v}$ is not $y$-good}
		\STATE $v^* \leftarrow \text{\textbf{SetGood}}(\hat{v})$
		\STATE $ \bw \leftarrow \bw + \phi(v^*) - \phi(\hat{v})$
	\ENDIF
	\ENDFOR
\ENDFOR
\end{algorithmic}
\end{algorithm}

The {\em speedup-learning} algorithm is listed as
Algorithm~\ref{speedup}. It begins by initializing the weight $\bw$ to
the zero vector. We iterate over the examples for $M$
epochs. 
For each example $\bx_i$, we first solve inference
using the ILP solver to obtain the reference structure $\by$ (line 4). Next a
breadth-expand search is performed (lines 5-8).
Every time the beam $B$ is updated, we check if the beam contains at
least one $y$-good node that can possibly lead to the reference solution
$\by$. Search terminates if the beam is not $y$-good, or if the highest ranking node $\hat{v}$ is a goal. If the beam is not $y$-good, we compute the corresponding
$y$-good node $v^*$ from $\hat{v}$, and perform a perceptron style update to the
speedup weights (line 9-11). In other words, we update the weight
vector by adding feature vector of $\phi(v^*)$, and subtracting the
average feature vector of all the nodes in the beam. Otherwise $\hat{v}$
must be a goal node. We then check if $\hat{v}$ agrees with the reference
solution (lines 12-15). If not, we perform a similar weight update, by
adding the feature vector of $\phi(v^*)$, and subtracting
$\phi(\hat{v})$.

\paragraph{Mistake bound}
Next, we show that the Algorithm~\ref{speedup} has a mistake
bound. Let $R_\phi$ be a positive constant such that for every
pair of nodes $(v, v')$, we have $\|\phi(v)-\phi(v')\|\le R_\phi$.
Let $R_g$ be a positive constant such that for every pair
of search nodes $(v, v')$, we have $|g(v)-g(v')|\le R_g$.
Finally we define the {\em level margin} of a weight vector $\bw$ for a
training set as 
\begin{equation}\label{level-margin}
\gamma = \min_{\{(v, v')\}} \bw \cdot \Big(\phi(v) - \phi(v')\Big)
\end{equation}
Here, the set $\{(v, v')\}$ contains any pair such that
$v$ is $y$-good, $v'$ is not $y$-good, and $v$ and $v'$ are at the
same search level. The level margin denotes the minimum score gap
between a $y$-good and a $y$-bad node at the same search level.

The priority function used to rank the search nodes is defined as $p_\bw(v) = g(v) - \bw \cdot \phi(v)$. Smaller priority function value ranks higher during search.
With these definitions we have the following theorem:
\begin{theorem}[Speedup mistake bound]
Given a training set such that there exists a weight vector $\bw$
with level margin $\gamma>0$ and $\|\bw\|=1$, the speedup learning
algorithm (Algorithm~\ref{speedup}) will converge with a consistent weight vector after making no more than $\frac{R_\phi^2+2R_g}{\gamma^2}$ weight updates.
\end{theorem}
\begin{proof}
The complete proof is in the supplementary material of the paper.
\end{proof}


\subsection{Avoiding Computing the Input Features}
\label{sec:theta-trick}
So far, we have shown that a structured prediction problem can be
converted to a beam search problem. The priority function for ranking
search nodes is determined by $p(v) = g(v) + h(v)$. We have seen how
the $h$ function be trained to enforce structural
constraints. However, there are other opportunities for speeding up as
well.

Computing the path cost $g(v)$ involves calculating the corresponding ILP coefficients, which in turn requires feature extraction using the original trained model. This is usually a time-consuming step~\cite{srikumar2017algebra}, thus motivating the question of whether we can avoid calculating them without losing accuracy.
If a search node is strongly preferred by the heuristic function, the path cost is unlikely to reverse the heuristic function's decision. In this case, we can rank the candidate search nodes with heuristic function only.

Formally, given a fixed beam size $b$ and the beam candidates $C_t$ at step $t$ from which we need to select the beam $B_t$, we can rank the nodes in $C_t$ from smallest to largest according to the heuristic function value $h(v)$. Denote the $b^{\text{th}}$ smallest node as $v_b$ and the $(b+1)^{\text{th}}$ smallest node as $v_{b+1}$, we define the heuristic gap $\Delta_t$ as
\begin{equation}
\Delta_t = h(v_{b+1}) - h(v_b).
\end{equation}
If the beam $B_t$ is selected from $C_t$ only according to heuristic function, then $\Delta_t$ is the gap between the last node in the beam and the first node outside the beam. Next we define the path-cost gap $\delta_t$ as
\begin{equation}
\delta_t = \max_{v,v' \in C_t} (v - v')
\end{equation}
With these definitions we immediately have the following theorem:
\begin{theorem}\label{h-only}
Given the beam candidates $C_t$ with heuristic gap $\Delta_t$ and
path-cost gap $\delta_t$, if $\Delta_t > \delta_t$, then using only
heuristic function to select the beam $B_t$ will have the same set
of nodes selected as using the full priority function up to their
ordering in the beam. 
\end{theorem}
If the condition of Theorem~\ref{h-only} holds, then we can rank the
candidates using only heuristic function without calculating the
path cost. This will further save computation time. However, without actually calculating the path cost there is no way to determine the path-cost gap $\delta_t$ at each step. In practice we can treat $\delta_t$ as an empirical parameter $\theta$ and define the following priority function
\begin{equation}\label{p-theta}
p_\theta(v) = 
\begin{cases}
h(v), & \text{if $\Delta_t>\theta$}, \\
g(v) + h(v), & \text{otherwise.}
\end{cases}
\end{equation}


\section{Experiments}
\label{sec:experiments}

We empirically evaluate the speedup based inference scheme described
in Section \ref{sec:learning} on the problem of predicting entities and
relations (i.e. our running example).  In this task, we are asked to
label each entity, and the relation between each pair of the
entities. We assume the entity candidates are given, either from human
annotators or from a preprocessing step. The goal of inference is to
determine the types of the entity spans, and the relations between
them, as opposed to identify entity candidates.
The research questions we seek to resolve empirically are:
\begin{enumerate}[nosep]
\item Does using a learned speedup heuristic recover structurally
  valid outputs without paying the inference cost of the integer
  linear program solver?
\item Can we construct accurate outputs without always computing input
  features and using only the learned heuristic to guide search?
\end{enumerate}

The dataset we used is from the previous work by \citet{Roth2004}. It contains 1441 sentences. Each sentence contains several entities with labels, and the labeled relations between every pair of entity. There are three types of entities, \lbl{person}, \lbl{location} and \lbl{organization}, and five types of relations, \lbl{Kill}, \lbl{LiveIn}, \lbl{WorkFor}, \lbl{LocatedAt} and \lbl{OrgBasedIn}. There are two constraints associated with each relation type, specifying the allowed source and target arguments. For example, if the relation label is \lbl{LiveIn}, the source entity must be \lbl{person} and the target entity must be \lbl{location}. There is also another kind of constraint which says for every pair of entities, they can not have a relation label in both directions between them, i.e., one of the direction must be labeled as \lbl{NoRel}. 

We re-implemented the model from the original work using the same set of features as for the entity and relation scoring functions. We used 70\% of the labeled data to train an ILP-based inference scheme, which will become our black-box solver for learning the speedup classifier. The remaining 30\% labeled data are held out for evaluations. 

We use 29950 sentences from the Gigaword corpus~\cite{graff2003english} to train the speedup classifier. The entity candidates are extracted using the Stanford Named Entity Recognizer~\cite{manning2014stanford}. We ignore the entity labels, however, since our task requires determining the type of the entities and relations.  The features we use for the speedup classifiers are counts of the pairs of labels of the form (source label, relation label), (relation label, target label), and counts of the triples of labels of the form (source label, relation label, target label). We run  Algorithm \ref{speedup} over this unlabeled dataset, and evaluate the resulting speedup classifier on the held out test set. In all of our speedup search implementations, we first assign labels to the entities from left to right, then the relations among them.

We evaluate the learned speedup classifier in terms of both accuracy
and speed. The accuracy of the speedup classifier can be evaluated
using three kinds of metrics: F-1 scores against gold labels, F-1
scores against the ILP solver's prediction, and the \emph{validity ratio}, which
is the percentage of the predicted examples agreeing with all
constraints.\footnote{All our experiments were conducted on a server
  with eight Intel i7 3.40 GHz cores and 16G memory. We disabled
  multi-threaded execution in all cases for a fair comparison.}

\subsection{Evaluation of Algorithm \ref{speedup}}

Our first set of experiments evaluates the impact of Algorithm \ref{speedup}. These results  are shown in Table~\ref{tab:without-theta}. We see the ILP solver achieves perfect entity and relation F-1 when compared with ILP model itself. It guarantees all constraints are satisfied. Its accuracy against gold label and its prediction time becomes the baselines of our speedup classifiers. We also provide two search baselines. The first search baseline just uses greedy search without any constraint considerations. In this setting each label is assigned independently, since the step cost of assigning a label to an entity or a relation variable depends only on the corresponding coefficients in the ILP objectives. In this case, a structured prediction problem becomes several independent multi-class classification problems.  The prediction time is faster than ILP but the validity ratio is rather low (0.29). The second search baseline is greedy search with constraint satisfaction. The constraints are guaranteed to be satisfied by using the standard arc-consistency search. The prediction takes much longer than the ILP solver (844 ms vs. 239 ms.).

We trained a speedup classifier with two different beam sizes. Even with beam width $b=1$, we are able to obtain $>95\%$ validity ratio, and the prediction time is much faster than the ILP model. Furthermore, we see that  the F-1 score evaluated against gold labels is only slightly worse than ILP model. With beam width $b=2$, we recover the ILP model accuracy when evaluated against gold labels. The prediction time is still much less than the ILP solver.

\begin{table*}[t!]
\begin{center}
\begin{tabular}{lrrrrrr}
\toprule
\bf Model          & \bf Ent. F1 Gold & \bf Rel. F1 Gold & \bf Ent. F1 ILP & \bf Rel. F1 ILP & \bf Time (ms) & \bf Validity Ratio \\ 
\midrule
ILP                & 0.827            & 0.482            & 1.000           & 1.000           & $239 \pm11.4$           & 1.00               \\ 
\midrule
search             & 0.822            & 0.334            & 0.877           & 0.546           & $170 \pm 6.5$          & 0.29               \\
constrained search & 0.800            & 0.572            & 0.880           & 0.741           & $844 \pm31.0$          & 1.00               \\
\midrule
speedup ($b=1$)    & 0.822            & 0.447            & 0.877           & 0.674           & $136 \pm 2.1$       & 0.96               \\
speedup ($b=2$)    & 0.844            & 0.484            & 0.930           & 0.752           & $158 \pm19.8$     & 0.95               \\
\midrule
\end{tabular}
\end{center}
\caption{\label{RY} Performance of the speedup classifier with different beam sizes, compared with the ILP solver and search without heuristics. CPU time is in milli-seconds, and averaged over five different runs with standard deviations.}
\label{tab:without-theta}
\end{table*}

\subsection{Experiments on Ignoring the Model Cost}
In this section, we empirically verify the idea that we do not always need to compute the path cost, if the heuristic gap $\Delta_t$ is large. We use the evaluation function $p_\theta(v)$ in Eq.~\eqref{p-theta} with different values of $\theta$ to rank the search nodes. The results are given in Table~\ref{with-theta}.

For both beam widths, $\theta =0$ is the case in which the original model is completely ignored. All the nodes are ranked using the speedup heuristic function only. Even though it has perfect validity ratio, the result is rather poor when evaluated on F-1 scores. When $\theta$ increases, the entity and relation F-1 scores quickly jump up, essentially getting back the same accuracy as the speedup classifiers in Table~\ref{tab:without-theta}. But the prediction time is lowered compared to the results from Table \ref{RY}.


\begin{table*}[t!]
\begin{center}
\begin{tabular}{lrrrrrrr}
\toprule
\bf Model       & $\bf \theta$ & \bf Ent. F1 Gold & \bf Rel. F1 Gold & \bf Ent. F1 ILP & \bf Rel. F1 ILP & \bf Time (ms) & \bf Validity Ratio \\ 
\midrule
speedup ($b=1$) & 0            & 0.21             & 0                & 0.173           & 0               & $39\pm 3.7$            & 1.00               \\
speedup ($b=1$) & 0.25         & 0.822            & 0.435            & 0.877           & 0.546           & $87\pm 2.5$            & 0.99               \\
speedup ($b=1$) & 0.5          & 0.822            & 0.455            & 0.877           & 0.672           & $114\pm 4.7$           & 0.98               \\
\midrule
speedup ($b=2$) & 0            & 0.373            & 0.152            & 0.377           & 0.139           & $55\pm 2.5$            & 1.00               \\
speedup ($b=2$) & 0.25         & 0.819            & 0.461            & 0.893           & 0.623           & $130\pm14.4$           & 0.99               \\
speedup ($b=2$) & 0.5          & 0.825            & 0.494            & 0.907           & 0.689           & $134\pm 4.0$           & 0.98               \\
\midrule
\end{tabular}
\end{center}
\caption{\label{RYtheta} Performance of the speedup classifier with different beam size and $\theta$ values. CPU Time is in milli-second, and averaged over five different runs with standard deviations.}
\label{with-theta}
\end{table*}


\section{Discussion and Related Work}
\label{sec:related}

The idea of learning memo functions to make computation more
efficient goes back to \citet{michie1968memo}. Speedup learning has
been studied since the eighties in the context of general problem
solving, where the goal is to learn a problem solver that becomes
faster as opposed to becoming more accurate as it sees more data.
\citet{fern2011speedup} gives a broad survey of this area.  In this
paper, we presented a variant of this idea that is more concretely
applied to structured output prediction.

Efficient inference is a central topic in structured prediction. In
order to achieve efficiency, various strategies are adopted in the
literature. Search based strategies are commonly used for this purpose
and several variants abound.  The idea of framing a structured
prediction problem as a search problem has been explored by several
previous
works~\cite{Collins2004,DaumeIII2005,Daume2009,Huang2012,Doppa2014}. It
usually admits incorporating arbitrary features more easily than
fully global structured prediction models like conditional random
fields~\cite{Lafferty2001}, structured perceptron~\cite{Collins2002},
and structured support vector
machines~\cite{Taskar2003a,Tsochantaridis2004}. In such cases too, inference can be
solved approximately using heuristic search. Either a fixed beam
size \cite{Xu2009a}, or a dynamically-sized beam \cite{Bodenstab2011}
can be used. In our work we fix the beam size. The key difference
from previous work is that our ranking function combines information
from the trained model with the heuristic function which
characterizes constraint information.
Closely related to the work described in
this paper are approaches that learn to prune the search
space~\cite{He2014,Vieira2016} and learn to select
features~\cite{Daum2013}.

Another line of recent related work focuses on discovering problem
level regularities across the inference space. These {\em amortized}
inference schemes are designed using deterministic rules for
discovering when a new inference problem can re-use previously
computed solutions~\cite{Srikumar2012a,Kundu2013} or in the context of
a Bayesian network by learning a stochastic inverse network that
generates outputs~~\cite{stuhlmuller2013learning}.

Our work is also related to the idea of imitation
learning~\cite{Daume2009,Ross2011,Ross2014,Chang2015}. In this
setting, we are given a reference policy, which may or may not be a
good policy. The goal of learning is to learn another policy to
imitate the given policy, or even learn a better one. Learning
usually proceeds in an online fashion. However, imitation learning
requires learning a new policy which is independent of the given
reference policy, since during test time the reference policy is no
longer available. In our case, we can think of the black-box solver
as a reference policy. During prediction we always have this solver
at our disposal, what we want is avoiding unnecessary calls to the
solver. Following recent successes in imitation learning, we expect
that we can replace the linear heuristic function with a deep
network to avoid feature design.

Also related is the idea of knowledge
distillation~\cite{bucilua2006model,hinton2015distilling,kim2016sequence},
that seeks to train a student classifier (usually a neural network) to
compress and mimic a larger teacher network, thus improve prediction
speed. The primary difference
with the speedup idea of this paper is that our goal is to be more
efficient at constructing internally self-consistent structures without explicitly searching over the combinatorially large output space with complex constraints. 


\section{Conclusions}
\label{sec:conclusions}

In this paper, we asked whether we can learn to make inference faster
over the lifetime of a structured output classifier. To address this
question, we developed a search-based strategy that learns to mimic a
black-box inference engine but is substantially faster. We further
extended this strategy by identifying cases where the learned search
algorithm can avoid expensive input feature extraction
to further improve speed without losing accuracy. We empirically
evaluated our proposed algorithms on the problem of extracting
entities and relations from text. Despite using an object-heavy
JVM-based implementation of search, we showed that by
exploiting regularities across the output space, we can outperform
the industrial strength Gurobi integer linear program solver in terms of
speed, while matching its accuracy.


\noindent\textbf{Acknowledgments}
We thank the Utah NLP group members and the
anonymous reviewers for their valuable feedback.


\bibliography{example_paper,cited}
\bibliographystyle{icml2018}

\newpage

\icmltitlerunning{Learning to Speed Up Structured Output Prediction (Supplementary Material)}

\setcounter{section}{0}
\setcounter{equation}{0}
\setcounter{theorem}{0}
\setcounter{@affiliationcounter}{1}

\twocolumn[
\icmltitle{Learning to Speed Up Structured Output Prediction \\
           (Supplementary Material)}



\icmlsetsymbol{equal}{*}

\begin{icmlauthorlist}
\icmlauthor{Xingyuan Pan}{ut}
\icmlauthor{Vivek Srikumar}{ut}
\end{icmlauthorlist}
\icmlaffiliation{ut}{School of Computing, University of Utah, Salt Lake City, Utah, USA}



\vskip 0.3in
]



\printAffiliationsAndNotice{}  

\section{Linear Heuristic Function}
In section 3.1 of the main paper we show that greedy search combined with the priority function $p(v)=g(v)+h^*(v)$ will lead to the exact solution of the original ILP (Eqs.~(3) to (5) in the main paper), under the condition that there is no duality gap. For convenience we repeat the definition of the optimal heuristic function $h^*(v)$ here:
\begin{equation}
h^*(v)= -\sum_{(k,i)\in v}\sum_j A^k_{ji} u^*_j(\bx) \label{h}
\end{equation}
In this section we show that the heuristic function in Eq.~\eqref{h} can be written as a linear function of the form $-\bw \cdot \phi(v)$. 

First, let us recap the meanings and ranges of indices $k$, $i$, and $j$ in Eq.~\eqref{h}:
\begin{itemize}
\item index $k$ (from $1$ to $K$): the $k^{\text{th}}$ categorical variable.
\item index $i$ (from $1$ to $n$): the $i^{\text{th}}$ label.
\item index $j$ (from $1$ to $m$): the $j^{\text{th}}$ constraint.
\end{itemize}

Second, recall that a search node $v$ is just a set
of pairs $\{(k, i)\}$, each element of which specifies that the
variable $y^k$ is assigned the $i^{\text{th}}$ label.

Now we can define a $K\times n\times m$ dimensional feature vector $\phi(v)$, the component of which is labeled by a tuple of indices $(k, i, j)$. Let the $(k, i,j)^{\text{th}}$ component of the feature vector be
\begin{equation}\label{phi}
\phi_{kij}(v) =
\begin{cases}
u^*_j(\bx), &\quad\text{if $(k, i) \in v$} \\
0, &\quad\text{otherwise}
\end{cases}
\end{equation}
Also define the corresponding weight parameter 
\begin{equation}\label{w}
w_{kij} = A^k_{ji}
\end{equation}
Clearly the heuristic function in Eq.~\eqref{h} is just $-\bw \cdot \phi(v)$, where $\phi$ and $\bw$ is defined in Eq.~\eqref{phi} and Eq.~\eqref{w}, respectively.

\section{Proof of Mistake Bound Theorem}
The goal of this section is to prove Theorem 1 in the main paper. We will prove two lemmas which will lead to the final proof of the theorem. Before introducing the two lemmas, we repeat the relevant definitions here for convenience.

Let $R_\phi$ be a constant such that for every
pair of search nodes $(v, v')$, $\|\phi(v)-\phi(v')\|\le R_\phi$.
Let $R_g$ be a constant such that for every pair
of search nodes $(v, v')$, $|g(v)-g(v')|\le R_g$.
Finally we define the {\em level margin} of a weight vector $\bw$ for a
training set as 
\begin{equation}\label{level-margin}
\gamma = \min_{\{(v, v')\}} \bw \cdot \Big(\phi(v) - \phi(v')\Big)
\end{equation} 
Here, the set $\{(v, v')\}$ contains any pair such that
$v$ is $y$-good, $v'$ is not $y$-good, and $v$ and $v'$ are at the
same search level. The priority functin used to rank the search nodes is defined as $p_\bw(v) = g(v) - \bw \cdot \phi(v)$. Smaller priority function value ranks higher during search.
With these definitions we have the following two lemmas:

\begin{lemma}\label{upperbound}
Let $\bw^k$ be the weights before the $k^{\text{th}}$ mistake is made ($\bw^1=\mathbf{0}$). Right after the $k^{\text{th}}$ mistake is made, the norm of the weight vector $\bw^{k+1}$ has the following upper bound:
\begin{equation*}
\| \bw^{k+1} \|^2 \le \| \bw^{k} \|^2 + R_\phi^2 + 2R_g
\end{equation*}
\end{lemma}
\begin{proof}
When the $k^{\text{th}}$ mistake is made, we get $\bw^{k+1}$ by using the update rule in either line 11 or line 14 of the algorithm. Let us consider the case of updating using line 11 first. We have
\begin{align}
\|\bw^{k+1}\|^2&=\|\bw^k + \phi(v^*) - \frac{1}{|B|}\sum_{v \in B} \phi(v) \|^2 \nonumber \\
&=\|\bw^k\|^2 + \|\phi(v^*) - \frac{1}{|B|}\sum_{v \in B} \phi(v) \|^2 \nonumber \\
&\quad + 2\bw^k \cdot \Big(\phi(v^*) - \frac{1}{|B|}\sum_{v \in B} \phi(v)\Big) \label{wk+1}
\end{align}
We will upper bound each term separately on the right side of Eq.~\eqref{wk+1}. To bound the second term we use the definition of $R_\phi$ and the properties of vector dot product:
\begin{align}\label{2nd}
&\|\phi(v^*) - \frac{1}{|B|}\sum_{v \in B} \phi(v) \|^2\nonumber \\
&=\frac{1}{|B|^2} \|\sum_{v\in B}(\phi(v^*)-\phi(v))\|^2 \nonumber \\
&=\frac{1}{|B|^2} \sum_{v\in B}(\phi(v^*)-\phi(v)) \cdot \sum_{v'\in B}(\phi(v^*)-\phi(v')) \nonumber \\
&=\frac{1}{|B|^2} \sum_{v, v' \in B}(\phi(v^*)-\phi(v)) \cdot (\phi(v^*)-\phi(v')) \nonumber \\
& \le \frac{1}{|B|^2} \sum_{v, v' \in B}\|\phi(v^*)-\phi(v)\|  \|\phi(v^*)-\phi(v')\| \nonumber \\
&\le \frac{1}{|B|^2} \sum_{v, v' \in B} R_\phi^2 \nonumber \\
&= R_\phi^2
\end{align}
To bound the third term on the right side of Eq.~\eqref{wk+1}, note that the update is happening because of the $k^{\text{th}}$ mistake is being made. Therefore for any node $v \in B$, we have (smaller priority function value ranks higher)
\begin{equation*}
p_{\bw^k}(v^*) \ge p_{\bw^k}(v)
\end{equation*}
Using the definitions of $p_{\bw^k}(v)$ and $R_g$:
\begin{equation*}
\bw^k \cdot \Big(\phi(v^*) - \phi(v)\Big) \le g(v^*) - g(v) \le R_g
\end{equation*}
Now we have the upper bound for the third term on the right side of Eq.~\eqref{wk+1}:
\begin{align}\label{3rd}
&2\bw^k \cdot \Big(\phi(v^*) - \frac{1}{|B|}\sum_{v \in B} \phi(v)\Big) \nonumber \\
&= \frac{2}{|B|}\bw^k \cdot \sum_{v \in B}\Big( \phi(v^*) - \phi(v)\Big) \nonumber \\
&= \frac{2}{|B|}\sum_{v \in B}\bw^k \cdot \Big( \phi(v^*) - \phi(v)\Big) \nonumber \\
&\le  \frac{2}{|B|}\sum_{v \in B} R_g \nonumber \\
&= 2R_g
\end{align}
Combining Eqs.~\eqref{wk+1}, \eqref{2nd} and \eqref{3rd} leads to:
\begin{equation*}
\|\bw^{k+1}\|^2 \le \|\bw^k\|^2 + R_\phi^2 + 2R_g
\end{equation*}

Next we consider the case of updating using line 14 of the algorithm, in which we have
\begin{align}
\|\bw^{k+1}\|^2&=\|\bw^k + \phi(v^*) - \phi(\hat{v}) \|^2 \nonumber \\
&=\|\bw^k\|^2 + \|\phi(v^*) - \phi(\hat{v}) \|^2 \nonumber \\
&\quad + 2\bw^k \cdot \Big(\phi(v^*) - \phi(\hat{v})\Big) \label{wk+1s}
\end{align}
Using the definition of $R_\phi$:
\begin{equation}\label{2nds}
\|\phi(v^*) - \phi(\hat{v}) \|^2 \le R_\phi^2.
\end{equation}
Also, since a mistake is made by the weight $\bw^k$, we know $p_{\bw^k}(v^*) \ge p_{\bw^k}(\hat{v})$, which implies
\begin{equation}\label{3rds}
\bw^k \cdot \Big(\phi(v^*) - \phi(\hat{v})\Big) \le g(v^*)- g(\hat{v}) \le R_g
\end{equation}
Combining Eqs~\eqref{wk+1s}, \eqref{2nds} and \eqref{3rds} again leads to 
\begin{equation*}
\|\bw^{k+1}\|^2 \le \|\bw^k\|^2 + R_\phi^2 + 2R_g
\end{equation*}
\end{proof}

\begin{lemma}\label{lowerbound}
Let $\bw^k$ be the weights before the $k^{\text{th}}$ mistake is made ($\bw^1=\mathbf{0}$). Let $\bw$ be a weight vector with level margin $\gamma$ as defined in Eq.~\eqref{level-margin}. Then
\begin{equation*}
\bw \cdot \bw^{k+1} \ge \bw \cdot \bw^k + \gamma
\end{equation*}
\end{lemma}
\begin{proof}
First consider the update rule of line 11 of the algorithm,
\begin{align}
\bw \cdot \bw^{k+1}  &= \bw \cdot \Big(\bw^k + \phi(v^*) - \frac{1}{|B|}\sum_{v \in B} \phi(v)\Big)\nonumber \\
&= \bw\cdot\bw^k + \frac{1}{|B|} \sum_{v\in B} \bw\cdot \Big(\phi(v^*) - \phi(v)\Big)\nonumber \\
&\ge \bw\cdot\bw^k + \frac{1}{|B|} \sum_{v\in B} \gamma \nonumber \\
&=\bw\cdot\bw^k +\gamma \nonumber,
\end{align}
where we use the definition of the level margin $\gamma$.
Next consider the update rule of line 14 of the algorithm,
\begin{align*}
\bw \cdot \bw^{k+1}  &= \bw \cdot \Big(\bw^k + \phi(v^*) -\phi(\hat{v})\Big) \\
&=\bw\cdot\bw^k + \bw\cdot \Big(\phi(v^*) -\phi(\hat{v})\Big) \\
&\ge \bw\cdot\bw^k +\gamma
\end{align*}
\end{proof}

Now we are ready to prove Theorem 1 of the main paper, which is
repeated here for convenience.
\begin{theorem}[Speedup mistake bound]\label{theorem}
Given a training set such that there exists a weight vector $\bw$
with level margin $\gamma>0$ and $\|\bw\|=1$, the speedup learning
algorithm (Algorithm~1) will converge with a consistent weight vector after making no more than $\frac{R_\phi^2+2R_g}{\gamma^2}$ weight updates.
\end{theorem}
\begin{proof}
Let $\bw^k$ be the weights before the $k^{\text{th}}$ mistake is made ($\bw^1=\mathbf{0}$). Using Lemma~\ref{upperbound} repetitively (induction on $k$) gives us
\begin{equation*}
\|\bw^{k+1}\|^2 \le k(R_\phi^2 + 2R_g)
\end{equation*}
Similarly, induction on $k$ using Lemma~\ref{lowerbound},
\begin{equation*}
\bw\cdot\bw^{k+1} \ge k\gamma.
\end{equation*}
Finally we have
\begin{equation*}
1 \ge \frac{\bw\cdot\bw^{k+1}}{\|\bw\| \|\bw^{k+1}\|} \ge \frac{k\gamma}{\sqrt{k(R_\phi^2 + 2R_g)}}
\end{equation*}
which gives us
\begin{equation*}
k \le \frac{R_\phi^2 + 2R_g}{\gamma^2}
\end{equation*}
\end{proof}

\section{Proof of Theorem 2 of the Main Paper}
In this section we prove Theorem 2 of the main paper. We repeat the relevant definitions and the theorem here for convenience.

Given a fixed beam size $b$ and the beam candidates $C_t$ at step $t$ from which we need to select the beam $B_t$, we can rank the nodes in $C_t$ from smallest to largest according to the heuristic function $h(v)$. Denote the $b^{\text{th}}$ smallest node as $v_b$ and the $(b+1)^{\text{th}}$ smallest node as $v_{b+1}$, we define the heuristic gap $\Delta_t$ as
\begin{equation}
\Delta_t = h(v_{b+1}) - h(v_b)
\end{equation}
If the beam $B_t$ is selected from $C_t$ only according to heuristic function, then $\Delta_t$ is the gap between the last node in the beam and the first node outside the beam. Next we define the path-cost gap $\delta_t$ as
\begin{equation}
\delta_t = \max_{v,v' \in C_t} (v - v')
\end{equation}
With these definitions we have the following theorem:
\begin{theorem}\label{h-only}
  Given the beam candidates $C_t$ with heuristic gap $\Delta_t$ and
path-cost gap $\delta_t$, if $\Delta_t > \delta_t$, then using only
heuristic function to select the beam $B_t$ will have the same set
of nodes selected as using the full priority function up to their
ordering in the beam. 
\end{theorem}
\begin{proof}
Let $v^* \in C_t$ be any node that is ranked within top-$b$ nodes by the heuristic function $h(\cdot)$, and let $v\in C_t$ be an arbitrary node that is \emph{not} within top-$b$ nodes. By definitions of $v_b$ and $v_{b+1}$ we have
\begin{equation*}
h(v^*) \le h(v_b) \le h(v_{b+1}) \le h(v)
\end{equation*}
Therefore,
\begin{equation}\label{hpart}
h(v)-h(v^*) \ge h(v_{b+1}) - h(v_b) = \Delta_t.
\end{equation}
Also by definition of $\delta_t$ we have
\begin{equation}\label{gpart}
g(v) - g(v^*) \ge -\delta_t
\end{equation}
Combining Eqs.~\eqref{hpart} and \eqref{gpart},
\begin{align*}
p(v) - p(v^*) &= \Big(g(v) + h(v)\Big) - \Big(g(v^*) + h(v^*)\Big) \\
&= \Big(h(v) - h(v^*)\Big) + \Big(g(v) - g(v^*)\Big) \\
& \ge  \Delta_t - \delta_t \\
& > 0.
\end{align*}
Thus for any node $v^*\in C_t$, if it is selected in the beam $B_t$ (ranked top $b$) by the heuristic funciton $h(\cdot)$, it will be selected in the beam $B_t$ by the full priority funciton $p(\cdot)$.
\end{proof}

\end{document}